\documentclass[10pt,a4paper,twoside,twocolumn, english]{IEEEtran}
\usepackage[latin1]{inputenc}
\usepackage{amsmath}
\usepackage{amsfonts}
\usepackage{epsfig}
\usepackage{graphicx}
\usepackage{physics}
\usepackage{algorithm}
\usepackage{algorithmic}
\usepackage{bm}
\usepackage{amssymb}
\usepackage{booktabs,colortbl,multirow}
\usepackage{subfig}
\usepackage[numbers,sort&compress]{natbib}
\newcommand{\RNum}[1]{\uppercase\expandafter{\romannumeral #1\relax}}
\onecolumn

\DeclareMathOperator*{\argmax}{argmax}

\author{Siqi~Nie, Meng~Zheng, Qiang~Ji
        \\Rensselaer Polytechnic Institute}

\title{Deep Regression Bayesian Network
	\\and Its Applications}
\begin{document}
\maketitle
\begin{abstract}
Deep directed generative models have attracted much attention recently due to their generative modeling nature and powerful data representation ability. In this paper, we review different structures of deep directed generative models and the learning and inference algorithms associated with the structures. We focus on a specific structure that consists of layers of Bayesian Networks due to the property of capturing inherent and rich dependencies among latent variables. The major difficulty of learning and inference with deep directed models with many latent variables is the intractable inference due to the dependencies among the latent variables and the exponential number of latent variable configurations.
Current solutions use variational methods often through an auxiliary network to approximate the posterior probability inference. In contrast, inference can also be  performed directly without using any auxiliary network to maximally preserve the dependencies among the latent variables.
Specifically, by exploiting the sparse representation with the latent space, max-max instead of max-sum operation can be used to overcome the exponential number of latent configurations. Furthermore, the max-max operation and augmented coordinate ascent are applied to both supervised and unsupervised learning as well as to various inference. Quantitative evaluations on benchmark datasets of different models are given for both data representation and feature learning tasks.
\end{abstract}


\section{Introduction}
Deep learning has become a major enabling technology for computer vision. By exploiting its multi-level representation and the availability of the big data, deep learning has led to dramatic performance improvements for certain tasks. Among different deep learning architectures, the Convolutional Neural Networks (CNNs) have achieved the most significant development and are being widely employed in computer vision in recent years. CNNs, however, are typically discriminative models and they are built mainly for discriminative tasks, such as classification. For generative modeling, the model needs capture the underlying data distribution as well as the mechanisms used to generate data,
including the uncertainties in the data and in the data generation process. CNNs are therefore not directly suitable for generative modeling.
The latest development in Generative Adversarial Nets (GANs)~\cite{goodfellow2014generative} can perform effective data generation. Based on
combining a discriminative CNN with a generative de-convolutional neural network (DCNN), GANs produce remarkable performance in generating realistic data. Since
GANs employ a simple standardized random vector to model data uncertainty; they remain mostly deterministic and cannot fully model the uncertainties in data.

\begin{figure}
	\begin{center}
		\includegraphics[width=0.8\linewidth]{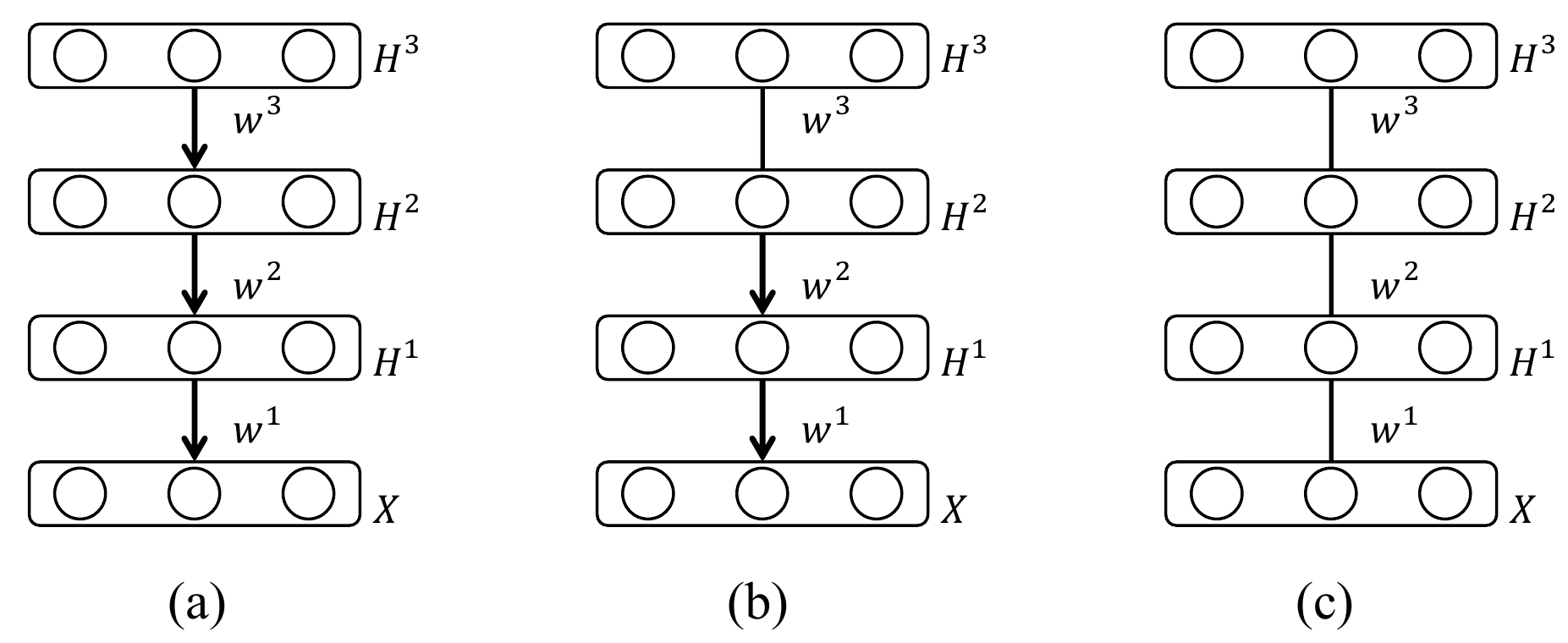}
	\end{center}
	\caption{Graph representation of (a) Deep Regression Bayesian Networks (DRBNs), (b) DBN, and (c) DBM,
where arrows represent directed links and line segments represent undirected links. Directed links capture the causal or one-way dependency between the connected nodes, while the undirected links capture the mutual dependencies among the connected nodes.}
	\label{fig:model}
\end{figure}

In this paper we focus on a fully probabilistic deep directed generative models for deep learning under uncertainty that can simultaneously perform data generation and classification tasks. Based on probability theories, deep probabilistic generative models offer a probabilistically grounded framework to represent, learn, and to predict under uncertainty. There exists several types of probabilistic deep generative models, most notably the Deep Boltzmann Machines (DBMs)~\cite{salakhutdinov2009deep} (Fig.~\ref{fig:model} (c)) and the Deep Belief Networks (DBNs)~\cite{hinton2006fast} (Fig.~\ref{fig:model} (b)). While generative in nature, these models, for the sake of simplicity in inference, typically assume latent variables are independent given data. Such an assumption weakens their data modeling and representation power.  In contrast, deep directed generative model consists of layers of Bayesian Networks as shown in (Fig.~\ref{fig:model} (a)).  Compared to DBMs and DBNs, the deep directed generative model enjoys several advantages due to its unique way in capturing dependencies. First, it specifically models the data generation process and allows straightforward ancestry sampling, where a node is sampled following a topological order, starting from its ancestors until its descendants. Second, there is no intractable partition function that has plagued the undirected models. Last but most importantly, latent nodes in the deep directed generative models are dependent on each other given inputs (because of the ``explaining away" principle) while it is not the case for DBN and DBM. This characteristic of  the deep directed generative models endows them powerful ability in modeling the complex pattern in data.


Learning and inference in deep directed generative models is challenging, mainly due to the intractable computation of the posterior probability of the latent variables and the exponential number of latent configurations (for binary latent variables). Various approximations such as variational inference algorithms \cite{rezende2014stochastic,Bachman2016,Han17aaai,kingma2013auto,mnih2014neural,goodfellow2014generative}
 have been proposed to address these challenges.
They all use an auxiliary feed forward network to approximately solve the intractable posterior probability inference problem.
 These approximations typically assume the joint latent variable distribution given data can be factorized. The factorized distribution, while simplifying the learning and inference, sacrifices the dependencies among the latent variables for efficiency. They inevitably enlarge the distance to the true posterior and weakens the representation power of the model. This negates a major advantage of the directed graphical models.
Moreover, the existing methods avoid the exponential number of latent configurations as they mostly deal with real latent nodes.

Alternatively, the posterior inference can be solved directly, without resorting to any other network. Furthermore, efficient learning and inference methods can be designed to maximally preserve the dependencies among the latent variables.  Specifically, for learning, data marginal log-likelihood can be maximized directly through a max-max operation to overcome the exponential number of latent configurations.  For inference, posterior probability inference can be performed through the pseudo-likelihood, which also preserves dependencies of latent variables. Furthermore, improved MAP inference can be achieved by combining coordinate ascent method with variational method.

The rest of the paper is structured as follows. In Section \RNum{2} we review related deep probabilistic generative models. In Section \RNum{3} we discuss the structure of a special kind of deep directed models, deep regression Bayesian networks (DRBNs). The learning and inference algorithms for DRBN are discussed in Section \RNum{4} and \RNum{5} respectively. Experimental evaluation of related models on data representation and classification is presented in Section \RNum{6} and \RNum{7}, and the paper is concluded in Section \RNum{8}.

\section{Related Work}
\label{sec:relat_work}
In general, there are three types of deep probabilistic generative models: fully directed, hybrid, and fully undirected. The DBN (Fig.~\ref{fig:model} (b)) has a hybrid structure, where the top two layers are connected with undirected links and all other layers are connected with directed links. The DBM (Fig.~\ref{fig:model} (c)) has a fully undirected structure such that every two consecutive layers are connected using undirected links. In this paper, we focus on the fully directed deep generative models due to their unique representation power. Depending on the structure and the type of variables, the directed generative models have several variants. The most commonly used structure consists of multiple layers (Fig.~\ref{fig:model} (a)), in which only the bottom layer represents the visible variables. The connections are directed from the upper layer to the lower layer, and no connection within each layer is allowed.

Depending on the types of variables, deep directed generative models can be categorized into deep sigmoid belief networks (SBNs)~\cite{neal1992connectionist,saul1996mean, mnih2014neural}, deep Poisson factor analysis (DPFA)~\cite{gan2015scalable}, deep factor analysers (DFAs)~\cite{tang2012deep}, and deep Gaussian mixture models (DGMMs)~\cite{van2014factoring}. The deep SBN contains binary latent and visible variables, and the conditional probability is defined using a sigmoid function. The DPFA models discrete variables (e.g., counts of words in documents) using binary latent variables. Dirichlet prior $\phi$ and gamma prior $\theta$ are placed to describe a Poisson distribution of input data. The DFA consists of continuous latent and mixtured types of variables. The DGMM is an extension of the Gaussian mixture model, while each latent node represents a linear operation to compute the mean and covariance matrix for the Gaussian distribution. 

In this paper, we focus on deep directed generative model with binary latent variables.  Compared to earlier SBN work \cite{neal1992connectionist, saul1996mean, mnih2014neural}, the deep directed generative models represent an
extension in both representation and learning and inference methods.  In terms of representation, it allows representing different input and output types (discrete, continuous, and hybrid), while SBN only uses binary input. For learning, DRBN allows both unsupervised and supervised layer-wise and global learning, while SBN work only includes layerwise unsupervised learning. For inference, DRBN allows of different algorithms such as the pseudo-likelihood method for posterior probability inference and augmented coordinate ascent method for MAP inference, while literatures for SBN generally use variational methods for inference.

For deep directed generative models, computing the posterior becomes intractable due to dependencies among the latent variables. To address this issue, one approach is to design a special prior to make the latent variables conditionally independent such as the complementary prior~\cite{hinton2006fast} for DBNs, wherein the posterior probability for each latent variable can be computed individually. Another popular approach is to replace the true posterior distribution with a factorized distribution as an approximation, known as variational methods. The mean field theory~\cite{saul1996mean} for learning sigmoid belief networks (SBNs) makes the latent variables totally independent. A set of variational parameters is learned to minimize the Kullback-Leibler (KL) divergence between the true posterior and the approximate one. Another approximate inference algorithm is the Markov chain Monte Carlo (MCMC) method, which can be used to estimate the posterior probability of both latent variables and parameters. One example is the learning and inference for deep latent Gaussian models~\cite{hoffman2017learning}.

To extend the traditional variational methods, recent works typically  use an auxiliary network to address the computational challenge with posterior probability inference. Specifically, the wake-sleep algorithm~\cite{hinton1995wake} augments the SBNs with a feed-forward recognition network for efficient inference.
Mnih and  Gregor \cite{mnih2014neural} propose a variational inference network with discrete latent variables to perform efficient inference.  Rezende et al.
\cite{rezende2014stochastic} and Kingma and Welling
\cite{kingma2013auto} introduced a recognition model to efficiently approximate posterior probability inference. The deep generative model can be learned by jointly optimization of the parameters for both generative and the recognition models. Ranganath et al. \cite{ranganath2014black} introduces a new method to reduce the variance of the noisy gradients for varational based deep model learning. The overall structure of \cite{Bachman2016} is similar to \cite{kingma2013auto}, by building connections between the recognition model and generative model rather than learning them independently.

All these approaches use a feed forward model to perform posterior probability inference. The feed forward models all assume the posterior probability can be factorized. The posterior probability inference is hence approximate.
In contrast, posterior probability inference can be done via coordinate ascent without the help of any auxiliary network or variational distribution, therefore during learning we do not need to assume conditional independence among the latent variables. Moreover, by initializing the coordinate ascent method with the inference result from a variational method, augmented coordinate ascent method can produce better reconstruction results than the existing variational models.

One of the latest development in deep generative modeling is the Generative Adversarial Nets (GAN) \cite{goodfellow2014generative,Han17aaai}, which employs both a discriminator and a generator. The generator is a deep generative model, which is trained to generate realistic synthetic data. The discriminator is a neural network trained to discriminate synthetic data from real data. The two networks compete against with each other until the discriminator is maximally confused. Specifically,
Han et al. \cite{Han17aaai} uses a nonlinear factor analyzer as the generator network. By performing the alternating back-propagation, the generator network can be learned to generate realistic images, sequences and sounds.  Despite its impressive generative performance, GANs are essentially a deterministic model as the data uncertainty is modeled by a standard random vector.  They hence cannot effectively capture the probabilistic distribution of the data.
 Furthermore, GANs can only perform data generation tasks and cannot perform classification tasks.
 By combining a variational autoencoder with a generative adversarial network, Larson et al.~\cite{larsen2016autoencoding} introduced a novel metric for similarity measurement when reconstructing the training samples during parameter updating, which achieves better generalization performance. Variational autoencoders and GANs are further combined in \cite{mescheder2017adversarial}, with a clear theoretical justification and the ability for arbitrary complex inference.

\section{Deep Regression Bayesian Networks}
\subsection{Regression Bayesian Networks}
\label{sec:RBN}
To construct a deep directed generative model, Bayesian Network (BN) is used as the building block.
A BN is parameterized by the conditional probability for each node, given its parents.  The number of parameters for each node
increases exponentially with the number of parents of each node. In order to scale up to models with a large number of latent variables, Regression BNs (RBNs) are employed to limit the number of parameters linearly with the number of connections~\cite{rijmen2008bayesian}.
\begin{figure}[ht]
	\begin{center}
		\includegraphics[width=0.8\linewidth]{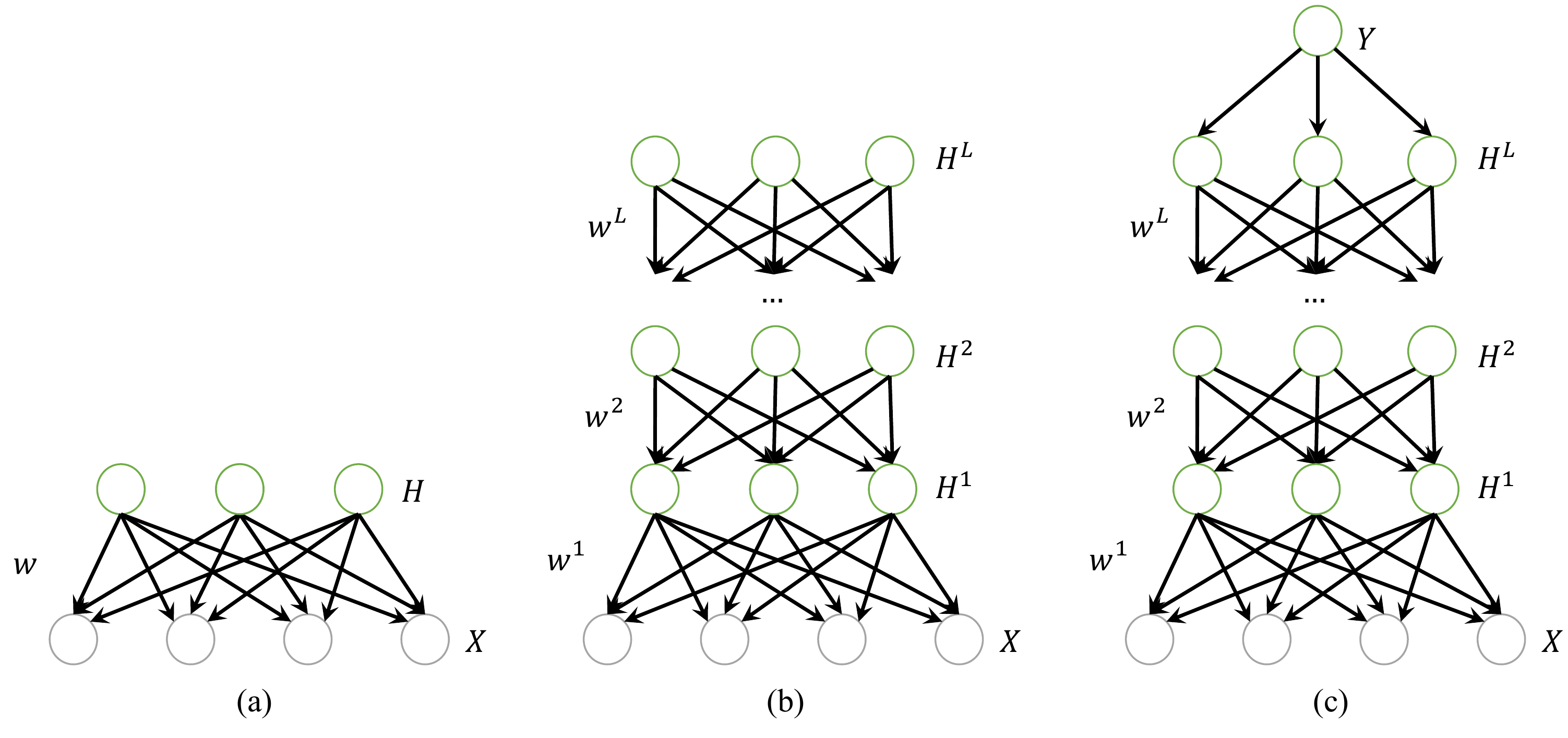}
	\end{center}
	\caption{Graph representation of (a) RBN, (b) DRBN without labels, and (c) DRBN with labels.}
	\label{fig:allrbns}
\end{figure}

The RBN consists of two layers: one visible layer $\bm{X}$ of dimension $n_d$ and one latent layer $\bm{H}$ of dimension $n_h$, as shown in Fig.~\ref{fig:allrbns} (a). Every latent variable connects to every visible variable with a directed edge.
For continuous data vector $\bm{X}=\{X_1,...,X_{n_d}\}\in \mathbb{R}^{n_d}$ and binary latent variables $\bm{H}=\{H_1,...,H_{n_h}\}\in \{0,1\}^{n_h}$ in which each node takes value $0$ or $1$, the prior and conditional probabilities of the model are defined as,
\begin{equation}
P(H_j=1) = \text{sigm}(d_j)\, ,
\label{prior}
\end{equation}
\begin{equation}
P\left(X_i=x_i|\bm{H}=\bm{h}\right)\sim\mathcal{N}\left(x_i| \bm{w}_i^T\bm{h} +b_i, \, \sigma_i^2\right)\, ,
\label{condiprob}
\end{equation}
where $\text{sigm}(a)=1/(1+\exp(-a))$ is the sigmoid function; $\mathcal{N}(a|\mu, \sigma^2)$ represents the Gaussian distribution with mean $\mu$ and variance $\sigma^2$.  $\bm{W}=[\bm{w}_1,...,\bm{w}_{n_d}]^T$ is the weight matrix, $w_{ij}$ is the weight of the link connecting nodes $H_j$ and $X_i$, where $i=\{1,2,...,n_d\}$, $j=\{1,2,...,n_h\}$. $\bm{d}=[d_1,...,d_{n_h}]^T$ and $\bm{b}=[b_1,...,b_{n_d}]^T$ are the bias parameters for $\bm{H}$ and $\bm{X}$ respectively, and $\bm{\sigma^2}=[\sigma_1^2,...,\sigma_{n_d}^2]^T$ are the conditional variances for $\bm{X}$. This generative model can be viewed as a diagonal Gaussian mixture model with the number of components exponential in the number of latent variables.
For binary data, the prior probability of the latent node remains the same and the conditional probability is,
\begin{equation}
P(X_i=1|\bm{H}=\bm{h})=\text{sigm}(\bm{w}_i^T\bm{h} +b_i)\, .
\label{cpt_binary}
\end{equation}

Through the parameterization in Eq. \ref{condiprob}, the mean of a Bernoulli node is the sigmoid function of linear combination of all nodes in the previous layer. RBN can be seen as a probabilistic generalization to Neural Network, where the value of nodes in the current layer is the nonlinear transformation (e.g. via a sigmoid activation) of the linear combination of all nodes in the previous layer. With the prior and conditional distributions, the joint distribution for $\bm{X}$ and $\bm{H}$ is,
\begin{equation}
P(\bm{x},\bm{h}|\Theta)=\prod_{j=1}^{n_h} P(h_j|\Theta_j)\prod_{i=1}^{n_d} P(x_i|\bm{h},\Theta_i)\, ,
\label{jointp}
\end{equation}
where $\Theta_i$ and $\Theta_j$ denote the parameters for $i$th visible node and $jth$ latent node. $\Theta=\{\bm{W},\bm{\sigma},\bm{b},\bm{d}\}$. Plugging the parameterization in Eq.~\eqref{prior} and Eq.~\eqref{condiprob} into Eq.~\eqref{jointp} yields,
\begin{equation}
P(\bm{x},\bm{h}|\Theta)=\frac{\exp(-E(\bm{x},\bm{h}|\Theta))}{(2\pi)^{n_d/2}\prod_i \sigma_i \prod_j\left(1+\exp(d_j)\right)}\, ,
\end{equation}
where
\begin{equation}
\begin{split}
E(\bm{x},\bm{h}|\Theta)=&\sum_i \frac{(x_i-b_i)^2}{2\sigma_i^2}-\sum_i \frac{x_i-b_i}{\sigma_i^2}\bm{w}_i^T\bm{h} \\
& -\bm{d}^T\bm{h}+\sum_i \frac{1}{2\sigma_i^2}(\bm{w}_i^T\bm{h})^2\, .
\end{split}
\label{jointenergy}
\end{equation}

Similarly, we can produce the joint probability for RBN with binary input by combining Eq.~\ref{cpt_binary} with Eq.~\ref{prior}.
The energy function in Eq.~\ref{jointenergy} is very similar to the energy function for the Gaussian-Bernoulli Restricted Boltzmann Machine (GRBM)~\cite{hinton2006reducing} in Eq. \ref{rbmenergy}.
\begin{equation}
\begin{split}
E_{\text{RBM}}(\bm{x},\bm{h})=\sum_i \frac{(x_i-b_i)^2}{2\sigma_i^2}-\sum_i \frac{x_i}{\sigma_i^2}\bm{w}_i\bm{h}-\bm{d}^T\bm{h}\, .
\end{split}
\label{rbmenergy}
\end{equation}
Comparing the two equations, it is clear that the energy function for RBN has an extra term (the last term of Eq. ~\eqref{jointenergy}). This extra term
explicitly captures the relationship among latent variables. This represents a major representation difference between RBNs and RBMs.

\subsection{Deep Regression Bayesian Networks}
By stacking multiple RBNs, we are able to construct a deep directed generative model with $L$ latent layers, called Deep Regression Bayesian Networks (DRBNs) (Fig.~\ref{fig:allrbns} (b)). Let $\bm{H}^l, l=1,\dots,L$ denote the binary latent variables in layer $l$, and $\bm{H}^0=\bm{X}$. Denote the parameters between layer $\bm{H}^{l-1}$ and $\bm{H}^l$ as $\Theta^l=\{\bm{W}^l,\bm{b}^{l}\}$. Top layer parameters are $\bm{d}$. The prior probability for a variable of the top layer is,
\begin{equation}
P(H_j^L=1) = \text{sigm}(d_j)\, .
\label{deepprior}
\end{equation}

The conditional probability for the remaining layers is,
\begin{equation}
P(H_j^{l-1}=1|\bm{H}^l=\bm{h}^l)=\text{sigm}(\bm{w}_j^{lT}\bm{h^l}+b_j^l)\, ,\quad \quad 2\leq l\leq L\, .
\end{equation}

The conditional probability $P(\bm{x}|\bm{h}^1)$ for the bottom two layers is the same as in Eq.~\eqref{condiprob} or Eq.~\eqref{cpt_binary}, depending on the type of the input data. The joint probability for all variables is,
\begin{equation}
P(\bm{x},\bm{h}^1,...,\bm{h}^L|\Theta)=P(\bm{x}|\bm{h}^1) P(\bm{h}^L) \prod_{l=2}^L P(\bm{h}^{l-1}|\bm{h}^l)\, .
\end{equation}
By fully capturing the joint probability distribution of input data and latent variables,  DRBN  can accurately capture three types of dependencies among different variables, namely dependencies among input variables, dependencies among hidden variables, and interactions between hidden and input variables.  Because of its explicit capture of these dependencies, compared with DBNs and DBMs, DRBNs can better capture the data distribution. Compared with the GANs, DRBNs explicitly capture the underlying probabilistic data distribution and can perform both data modeling and prediction tasks.

\subsection{Comparison with Deep Neural Networks}
Compared with Deep Neural Networks (DNNs), DRBNs consist of layers of Regression BNs. Each hidden layer of BN hence captures the distribution of the input data at  different levels. In contrast, DNNs consist of layer of perceptions and each hidden layer summarizes the sufficient statistics (such as the mean) of the input at different levels. Hence, the main difference between DRBNs and DNNs is that DRBN captures the probabilistic distribution of the data, while DNN captures mean statistics of the data. Based on this understanding, DRBN represents a generalization to DNN and DRBN becomes DNN if the conditional variance for each latent node becomes zero. DRBN is therefore more powerful in data representation, in particular in representing the uncertainties in the data. On the other hand, the power in representation also leads to challenges in DRBN learning and inference and in its ability to scale up. DNNs are much better than DRBNs in efficient inference and learning, and in scaling up to large model. This explains why so far DNNs remain the dominant deep model architecture. But the promise of DRBN in data representation deserves further research to address its challenges.

\section{DRBN Learning}
\label{sec:learning}
\subsection{Unsupervised Learning}
We first discuss the unsupervised RBN learning and then extend it to DRBN learning. The goal of unsupervised parameter learning for an RBN is to estimate the parameters $\Theta$ given a set of data samples $\mathcal{D}=\{\bm{x}^{m}\}_{m=1}^M$. To maximally preserve the dependencies among the latent variables, it would be better to directly maximize the marginal log-likelihood instead of its surrogate (such as its lower bound), i.e.,
\begin{equation}
\Theta^* = \argmax_{\Theta}\log P(\mathcal{D}|\Theta) =\argmax_{\Theta} \sum_m \log \sum_{\bm{h}}P(\bm{x}^{m},\bm{h}|\Theta)\, .
\label{obj}
\end{equation}
Maximizing the marginal likelihood has two computational challenges: (1) computing the posterior probability $P(\bm{h}|\bm{x})$ is intractable even for one configuration $\bm{h}$ due to the dependencies among elements of $\bm{h}$; (2) there are exponential number of terms in the summation over $\bm{h}$.
To address these two challenges, $\Theta$ can be estimated by replacing the max-sum operation in Eq. \ref{obj} with
the max-max operation below, i.e.,
\begin{equation}
\Theta^* = \argmax_{\Theta} \log P(\mathcal{D}|\Theta) \approx\argmax_{\Theta} \sum_m  \log \max_{\bm{h}} P(\bm{x}^{m},\bm{h}|\Theta)\, .
\label{max_obj}
\end{equation}

The rational for replacing the max-sum with max-max operation is based on the empirical observation that the distribution of $P(\bm{h}| \bm{x}^m)$ is very sparse, with its energy concentrated on a few configurations of $\bm{h}$ near its maximum. The max-max operation may be achieved iteratively in two steps. First, for each training sample $\bm{x}^m$,  we obtain $\bm{h}^{*m}$ that maximizes $P(\bm{h}, \bm{x}^m|\Theta_{t-1})$, given current parameters $\Theta_{t-1}$, i.e.,
\begin{eqnarray}
\bm{h}^{*m}=\argmax_{\bm{h}} P(\bm{h},\bm{x}^m|\Theta_{t-1})
\label{max_h}
\end{eqnarray}
Second, given $\bm{h}^{*m}$, $\Theta_t$ can be estimated by maximizing $\sum_{m} \log P(\bm{x},\bm{h^*}_m|\Theta_t)$, i.e.,
\begin{eqnarray}
\Theta^*_t=\arg\max_{\Theta} \log \sum_{m} P(\bm{x}^m, \bm{h}^{*m}|\Theta_t)
\label{max_likelihood}
\end{eqnarray}
The two steps iterate until convergence.  Eq. \ref{max_h} can be solved through the augmented coordinate ascent method to be
discussed in section~\ref{ACA}.  Eq. \ref{max_likelihood} can be solved in closed form solution for continuous input or  through stochastic gradient ascent for binary input.

Learning a DRBN with multiple latent layers consists of two steps: layerwise pre-training and global fine-tuning. Layerwise pre-training is a bottom-up procedure. When learning the parameters $\Theta^l$ for the $l$th layer,  the parameters below are frozen, and
the input to the $l$th layer is $\bm{H}^{*l}$ , which is obtained though $\bm{H}^{*l}=\arg\max_{\bm{H}^l}P(\bm{H}^l|\bm{H}^{l-1},\Theta^{l-1})$.
Given its input, $\Theta^l$ can be learned the same way using the RBN learning method.
The layerwise learning does not consider the interactions among the layers.  A global fine-tuning can then be performed.
Initialized by the parameters learnt by layer-wise learning, the global fine tuning procedure simultaneously refines the parameters
at different layers such that parameters in the higher layers can influence those in the lower layers. Specifically, global fine tuning updates the parameters $\bm\Theta$ in all layers by maximizing the marginal likelihood of the data, i.e.,
\begin{eqnarray}
\begin{split}
\bm\Theta^* = &\arg\max_\Theta \sum_{m} \log P(\bm{x}^{m}|\bm\Theta) \nonumber \\
 \approx &\arg\max_\Theta \sum_{m} \log \max_{\bm{h}^1,...,\bm{h}^L}P(\bm{x}^{m},\bm{h}^1,...,\bm{h}^L|\bm\Theta)\, .
\end{split}
\label{global_fine_tuning}
\end{eqnarray}
Eq. \ref{global_fine_tuning} can be solved through two step iteration as for Eq. \ref{max_obj}.

\subsection{Supervised Learning}
For applications where labels $\bm{y}^m$ is given for each sample $\bm{x}^m$ as shown in Fig. \ref{fig:model}, supervised learning can be performed. For classification tasks, discriminative supervised learning can be performed. The objective function is modified to maximize the log posterior probability of the labels,
\begin{equation}
\begin{split}
\bm\Theta^*=&\argmax_{\Theta} \log P(\mathcal{Y}|\mathcal{D},\bm\Theta) \\
= &\argmax_{\Theta}\sum_m \log P (y^{m}|\bm{x}^{m},\bm\Theta) \\
= &\argmax_{\Theta}\sum_m \log \sum_{\bm{h}} P (y^{m}, \bm{h}|\bm{x}^{m},\bm\Theta) \, .
\end{split}
\label{dis_learning}
\end{equation}
We encounter the same computational challenges in solving Eq. \ref{dis_learning}.  This challenge can be alleviated by replacing the max-sum operation with the max-max operation (Eq. \ref{dis_learning_max}),i.e., which can be solved similarly through the two-step iteration process.
\begin{equation}
\bm\Theta^*= \argmax_{\Theta}\sum_m \log \max_{\bm{h}} P (y^{m}, \bm{h}|\bm{x}^{m},\bm\Theta)\, .
\label{dis_learning_max}
\end{equation}

Like unsupervised learning,  Eq. \ref{dis_learning_max} can be applied to both layerwise and global DRBN learning. In order to achieve a good initialization of parameters, the generative model is first trained  in an unsupervised manner. The objective function is then changed to the supervised one, and parameters are tuned in a supervised manner.


\section{DRBN Inference}
Given a DRBN with known parameters, there are three types of inferences: posterior probability inference, MAP inference, and likelihood inference. In this section, we discuss efficient methods for the three types of inferences.

\subsection{Posterior Probability Inference}
The posterior inference is to compute the posterior probability of the latent variables given the input data, i.e., to compute $P(\bm{h}|\bm{x})$.
Because of the dependencies among $\bm{h}$, directly computing $P(\bm{h}|\bm{x},\Theta)$ is computationally intractable when the dimension of $\bm{h}$ is high. The pseudo-likelihood method offers an efficient solution to this problem by replacing the conditional likelihood with a more tractable objective. The pseudo-likelihood method considers the following approximation,
\begin{equation}
P(\bm{h}|\bm{x},\Theta)\approx \prod_j P(h_j|\bm{h}_{-j},\bm{x},\Theta)\, ,
\label{pseudo-likelihood}
\end{equation}
where $\bm{h}_{-j}=\{h_1,\dots,h_{j-1},h_{j+1},\dots,h_{n_h}\}$. In this approximation, conditioning is added over additional variables. The conditional pseudo-likelihood can be factorized into local conditional probabilities, which can be computed in parallel.  The pseudo-likelihood approximation, however, requires an initialization of $\bm{h}$.

\subsection{MAP Inference}
\label{ACA}
MAP inference is to estimate the most likely hidden layer configuration $\bm{h}^*$, given an input $\bm{x}$, i.e., $\bm{h}^*=\arg\max_{\bm{h}}P(\bm{h}|\bm{x})$. Directly performing MAP inference is computationally intractable as we need
enumerate all possible hidden layer configurations. Coordinate ascent method was introduced to overcome this challenge, by
iteratively maximizing one latent variable at a time. From an initial state of the latent vector $\bm{h}^{(0)}$, the coordinate ascent method updates one latent variable by fixing all other variables iteratively,
\begin{equation}
h_j^{(t+1)}=\argmax_{h_j} P(h_j|\bm{x}, \bm{h}^{(t)}_{-j})\, .
\label{map_ca}
\end{equation}

This iterative updating rule guarantees that the posterior probability $P(\bm{h}|\bm{x})$ will only increase until convergence due to the inequality:  $P(h_j^{t+1},\bm{h}^t_{-j}|\bm{x})\geq P(\bm{h}^t|\bm{x})$. As a greedy approach, the coordinate ascent method may get stuck in a local optimum. Thus, the initialization for Eq. \ref{map_ca} is crucial to ensure a configuration with high quality. To address this issue, the variational inference approach may be employed to learn an inference network~\cite{mnih2014neural} from the DRBN. The inference result from the inference network is used as initialization for the coordinate ascent, yielding the augmented coordinate ascent method.

The inference network method~\cite{mnih2014neural} approximates the posterior using only one set of parameters for all data samples by defining a feed-forward network for $Q(\bm{h}|\bm{x},\bm{\phi})$. Inference network also assumes independencies among latent variables given input data,
\begin{equation}
Q(\bm{h}|\bm{x},\bm{\phi})=\prod_j q(h_j|\bm{x},\phi_j)\, .
\label{eq:in}
\end{equation}

Each individual probability is defined using a sigmoid function,
\begin{equation}
q(h_j|\bm{x},\phi_j)=\text{sigm}\left(\sum_i v_{ij}x_i+s_j\right)\, .
\end{equation}

The parameters $\bm{\phi}=\{\bm{v}, \bm{s}\}$ are learned by minimizing the average KL divergence $KL(Q(\bm{h}|\bm{x},\bm{\phi})||P(\bm{h}|\bm{x},\Theta))$ over all the training samples. Similarly, the optimization is through the expected log-likelihood due to the intractable $P(\bm{h}|\bm{x},\Theta)$,
\begin{equation}
\bm{\phi}^*=\argmax_{\bm{\phi}} \sum_{\bm{x}} \sum_{\bm{h}} Q(\bm{h}|\bm{x},\bm{\phi})\log \frac{P(\bm{h},\bm{x}|\Theta)}{Q(\bm{h}|\bm{x},\bm{\phi})}\, .
\end{equation}

The coordinate ascent method requires to compute $P(h^t_j=1|\bm{x},\bm{h}^{t-1}_{-j})$ for each $j$. Naive computation can become prohibitive, given a large number of hidden variables. Taking discrete input data as illustration, according to Eq. \ref{jointp}, the computation of joint probability requires $N_h + N_d$ times of multiplication of the exponential terms. This can become very expensive when $N_d$ and $N_d$ are large. Since each time, only one element of $\bm{h}$ is changed and the probabilities for other elements remain the same. An efficient procedure to recursively compute $P(h^t_j=1|\bm{x},\bm{h}^{t-1}_{-j})$ can be written as,
\begin{equation}
\centering
\begin{split}
&P(h_j^t = 1|\bm{x},\bm{h}_{-j}^{t-1})\\
 = &\frac{P(h_j^t = 1,\bm{x},\bm{h}^{t-1}_{-j})}{P(h_j^t = 1,\bm{x},\bm{h}_{-j}^{t-1}) + P(h_j^t = 0,\bm{x},\bm{h}_{-j}^{t-1})}\\
= &\frac{1}{1+\frac{P(h_j^t = 1,\bm{x},\bm{h}_{-j}^{t-1})}{P(h_j^t = 0,\bm{x},\bm{h}_{-j}^{t-1})}} \, ,\\
\label{eq:GibbsRatio}
\end{split}
\end{equation}
where the probability ratio can be further written as,
\begin{equation}
\begin{split}
& \frac{P(h_j^t = 1,\bm{x},\bm{h}_{-j}^{t-1})}{P(h_j^t = 0,\bm{x},\bm{h}_{-j}^{t-1})} = \exp(d_j) \exp(x_i \bm{w}_j) \cdot \\
&\prod_{i} \frac{1+\exp(\sum_{n=1,n\neq j}^{n_h}w_{i,n} h_n^{t-1} + w_{i,j}+ b_i)}{1+\exp(\sum_{n=1,n\neq j}^{n_h}w_{i,n} h_n^{t-1}  + b_i)} \, .
\end{split}
\label{eq:GibbsSimple}
\end{equation}

The first two terms in Eq. \ref{eq:GibbsSimple} are constant for a given $\bm{x}$.  They need be computed only once. For the last term, either numerator or denominator can be retrieved from the last iteration, depending on the last value of $h^{t-1}_j$. This means we only need compute either the numerator or denominator in each iteration. This cuts the computational time by half compared to the naive computation. Furthermore, the term within the exponential term can be retrieved and updated from last iteration by one addition or subtraction. Then the probability ratio in Eq. \ref{eq:GibbsSimple} can be obtained by $n_d$ times of multiplication. As a result, computation time for each iteration is constant, involving one addition or subtraction and $n_d$ times of multiplication. The overall complexity of the coordinate ascent method is $O(n_d n_h n_i)$, which is linear in the size of the input data, the number of latent variables $n_h$, and the number of iterations $n_i$.

\subsection{Likelihood Inference}
The likelihood inference is to compute the marginal probability of the visible variables, i.e., $P(\bm{x})$.
Since
\begin{equation}
P(\bm{x})=\sum_{\bm{h}}P(\bm{h}|\bm{x})P(\bm{x})\, .
\end{equation}

By replacing the sum operation above with the max operation, we have
\begin{equation}
P(\bm{x})\approx \max_{\bm{h}}P(\bm{h}| \bm{x})P(\bm{x}) =\max_{\bm{h}}P(\bm{h}, \bm{x}) \, .
\label{eq:likelihood}
\end{equation}

It again can be computed by first computing $\bm{h^*}=\argmax_{\bm{h}} P(\bm{h}|\bm{x})$ using the augmented coordinate ascent method.
This can then be followed by computing $P(\bm{x},\bm{h}^*)$, which can be easily computed.

\section{Algorithm Evaluation}
\label{alg_evaluation}
As discussed in Section \ref{sec:relat_work}, DRBN's learning method is different from variational methods, which use a factorized variational distribution to approximate the true posterior probability inference. In contrast, DRBN directly maximizes the marginal log-likelihood through a max-max operation
To demonstrate the advantage of such a learning algorithm, we performed experiments to compare three learning methods: max-max learning, variational learning~\cite{mnih2014neural}, and the exact learning method in terms of
their ability in data representation under different number of latent nodes.

For this experiment, we trained the RBN networks on 60,000 training samples of MNIST dataset \cite{mnistdataset} (binarized according to~\cite{murray2009evaluating}). During learning, we respectively use the max-max operation, the variational method \cite{mnih2014neural}, and the exact method (only for small network) to approximate the marginal log-likelihood.  For a fair comparison, the hyper-parameters for each method are optimally tuned.

After learning, the mean log-likelihood of all training samples for RBNs learned with the three methods are summarized in Table \ref{small_loglike}.
It is clear from this table that
 the max-max method produces an average log-likelihood very close to the exact method, and is much better than the variational method. This demonstrates the improved accuracy
 of the max-max method in data representation over the variational methods for a small network.


\begin{table}
	\centering
	\caption{Comparison of training data log-likelihood (small network with 5 hidden nodes). }
	\scalebox{1.2}{
		\begin{tabular}{  ccc }
			\toprule
		   Variational \cite{mnih2014neural} & Max-max & Exact \\
			\midrule
			 -206.0620 & -170.7570 & -170.5416 \\
			\bottomrule
		\end{tabular}
	}
	\label{small_loglike}
\end{table}

\begin{figure}[h]
	\centering
	\includegraphics[width=0.8\textwidth]{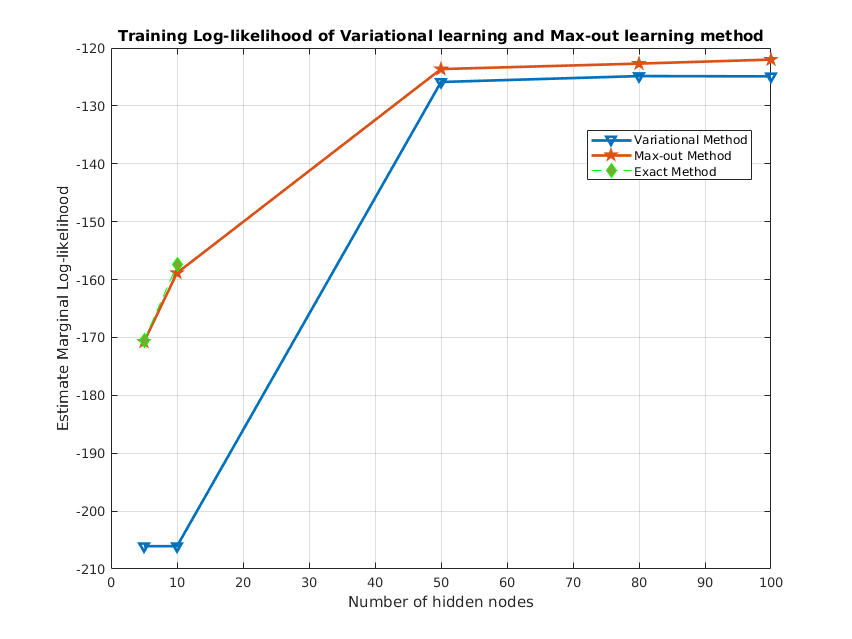}
	\caption{The average log-likelihood for RBN learned with different learning methods under different number of hidden nodes. Red line: max-out learning, blue line: variational learning, green line: exact learning.}
	\label{fig:algoeval}
\end{figure}

We further evaluate these methods for larger networks,  For network with a large number of hidden nodes, the exact method cannot be completed as the summation over
$\bm{h}$ becomes computationally intractable. We hence only give the mean data log-likelihood in Fig. \ref{fig:algoeval} for the max-max method and the variational method under  different number of hidden nodes. The estimate mean data log-likelihood is computed the same way as in Table \ref{small_loglike}.
We also include the mean log-likelihood of exact learning method for small RBN with 5 and 10 hidden nodes.
From the curves in this figure, we can see that data log-likelihood evaluated on RBN learned with max-max method is consistently higher than the variational method, especially when the network is small. This not only further demonstrates the advantage of the max-max learning method over the variational learning method but also proves the validity of replacing max-sum operation by the max-max operation.

\section{Applications}
To demonstrate the effectiveness of the DRBN models, we will apply DRBN to different computer vision tasks to demonstrate its capability for both data representation and feature learning for classification.

\subsection{Data Representation}
\label{sec:A}
First, we quantitatively compare DRBN with other generative models in terms of peak signal-to-noise ratio (PSNR) in several image restoration and denoising tasks. Then, we evaluate the generative representation power of different models in terms of generating synthetic data.

The first task is image restoration, which is to restore an image from its corrupted version. It is shown that a higher likelihood of patches leads to better denoising on whole images~\cite{zoran2011learning}. Therefore, we train a DRBN model on the 8$\times$8 patches from the Berkeley data set~\cite{martin2001database}, which contains 200 training and 100 testing images. One million training and 50,000 testing patches are randomly sampled from 200 training and 100 test images, respectively. A DRBN with two latent layers is used, with each layer containing 50 latent variables respectively. The total training epochs are 100. With a learnt DRBN as prior model, we use the Expected Patch Log Likelihood (EPLL)~\cite{zoran2011learning} framework to perform image restoration. The EPLL of an image $\bm{x}$ is defined as,
\begin{equation}
EPLL_P(\bm{x})=\sum_i \log P(\bm{y}_i)\, ,
\end{equation}
where $\{\bm{y}_i\}$ represents all the overlapping patches in the image. Given a corrupted image $\tilde{\bm{x}}$, the cost we use in order to reconstruct the image with patch prior $P$ is,
\begin{equation}
f_P(\bm{x}|\tilde{\bm{x}})=\frac{\lambda}{2} \|\bm{x}-\tilde{\bm{x}}\|^2 - EPLL_P(\bm{x})\, .
\end{equation}

It is difficult to directly optimize the complicated cost function. We employ the Half Quadratic Slitting method~\cite{geman1995nonlinear} with $\lambda$ set to $10^6$, following the settings in~\cite{zoran2011learning}. To perform image inpainting, we superimposed some sentences on the clean image as the noise.  During optimization, both the coordinate ascent method (CA) and the augmented coordinate ascent method (AugCA) were used to perform MAP inference for each patch. Typically the CA and AugCA converge after 3 iterations, and initialization affects the final configuration of latent variables but not posterior likelihood. We compared DRBN to three state-of-the-art approaches with generic priors: Field of Experts (FoE)~\cite{roth2005fields}, KSVDG~\cite{elad2006image}, and the Gaussian mixture model~\cite{zoran2011learning} with full covariance matrix.  The quantitative results on the 100 test images are given in Table \ref{inpaintpsnr}.
DRBN
outperforms all other approaches in terms of PSNR values.

\begin{figure}[h]
	\centering
	\subfloat[]
	{
		\includegraphics[width=0.4\textwidth]{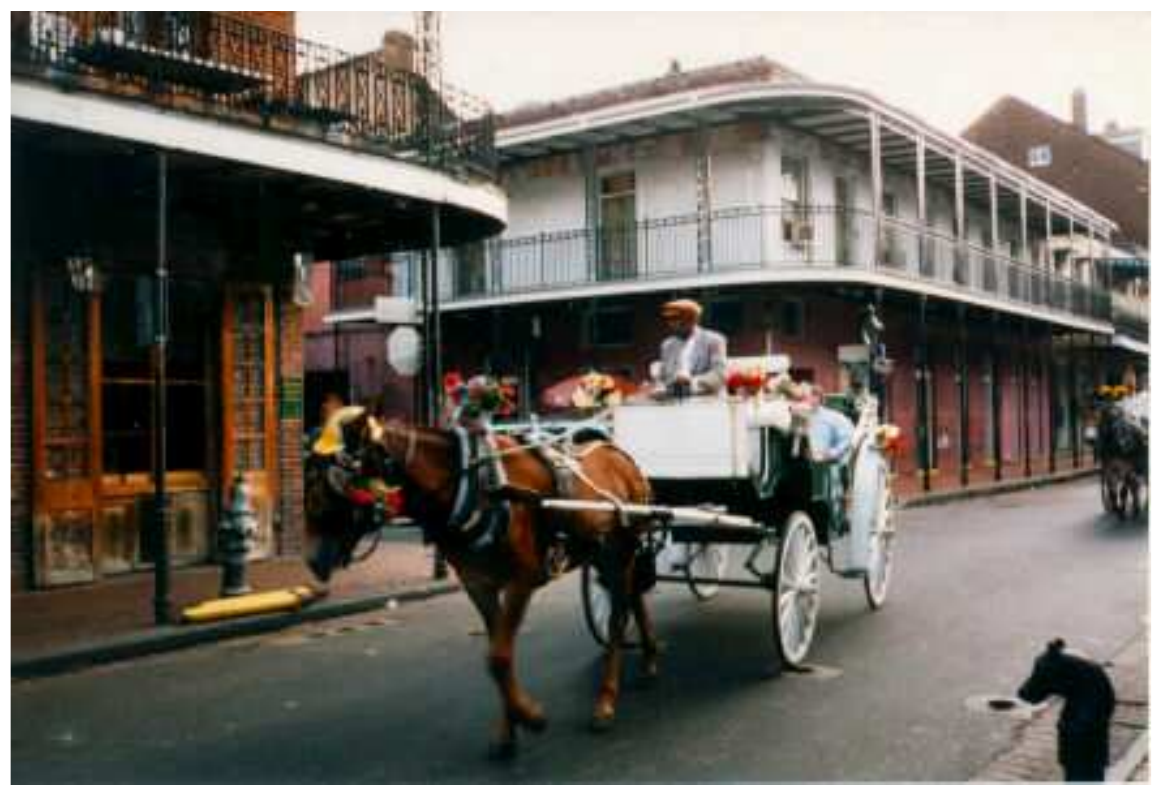} }
	\subfloat[]
	{
		\includegraphics[width=0.4\textwidth]{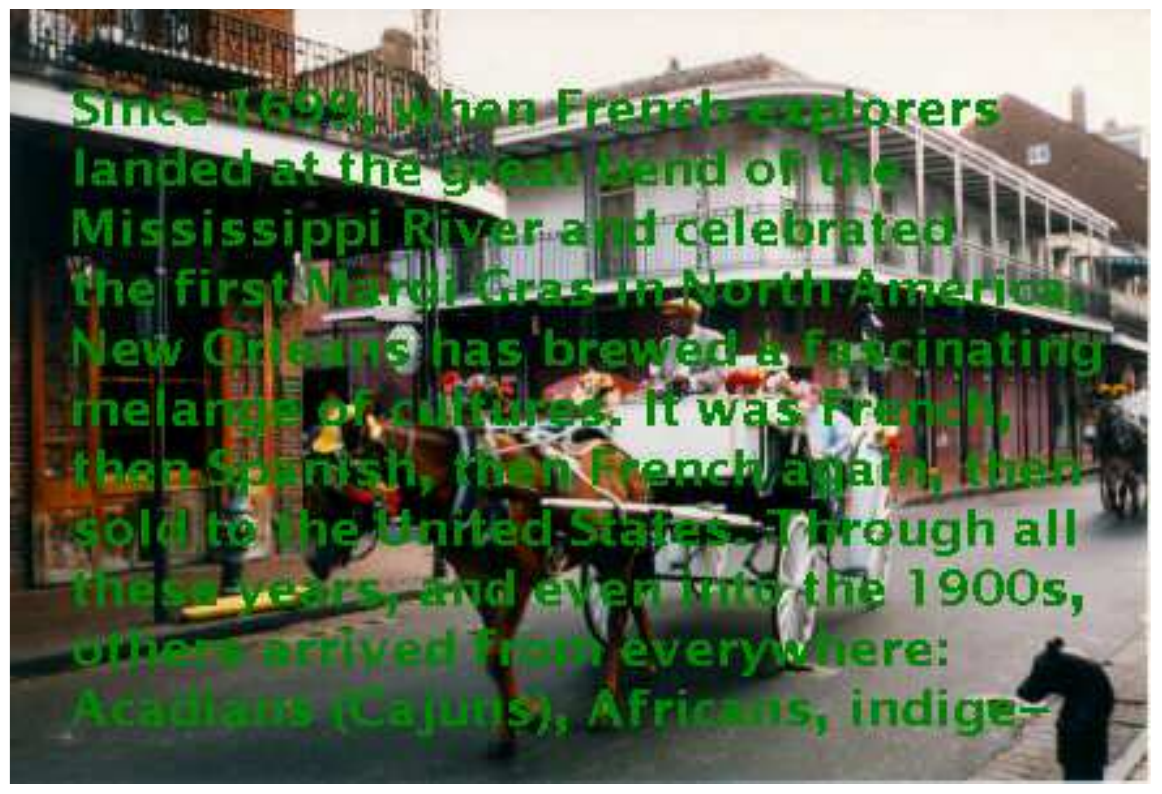} } \\
	\subfloat[]
	{
		\includegraphics[width=0.4\textwidth]{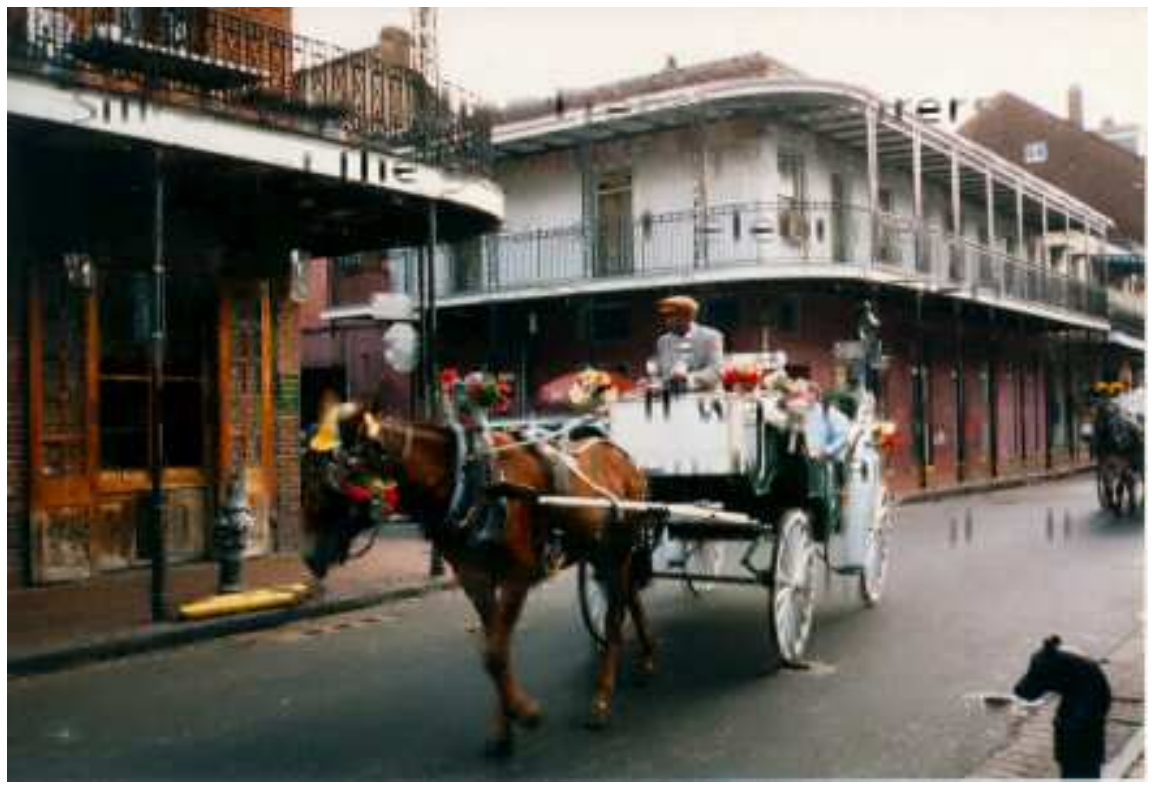} }
	\subfloat[]
	{
	\includegraphics[width=0.4\textwidth]{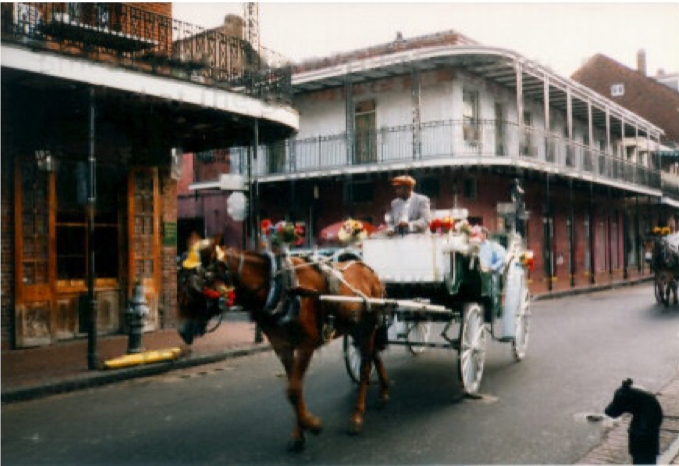} }
	\caption{An example of the image inpainting experiment, (a) original image, (b) corrupted image, (c) GMM, and (d) DRBN. Images collected from \cite{martin2001database}}
	\label{fig:inpaint}
\end{figure}

\begin{table}
	\centering
	\caption{PSNRs for different image inpainting methods on Berkeley data set. }
	\scalebox{1.2}{
		\begin{tabular}{lc}
			\toprule
			Method							& PSNR \\
			\midrule
			KSVD~\cite{elad2006image} 		& 24.13 \\
			FoE~\cite{roth2005fields} 		& 24.79 \\
			GMM~\cite{zoran2011learning} 	& 25.71 \\
			\midrule
			DRBN (CA) 				& 29.42 \\
			DRBN (AugCA)  			&33.46 \\
			\bottomrule
		\end{tabular}
	}
	\label{inpaintpsnr}
\end{table}
An example is given in Fig.~\ref{fig:inpaint}. The images represent an (a) original image, (b) corrupted image, (c) restored image using GMM as prior model, (d) restored image using DRBN as prior model. The PSNR values for (c) and (d) are 26.31 and 29.80, respectively. It can be seen that using the GMM prior, some parts of the letters in the corrupted image remain in the restored image, especially if the background color is white.  DRBN model completely removes the letters, and the PSNR values show significant improvement.

\begin{figure}
	\centering
	\subfloat[]
	{
	\includegraphics[width=0.5\textwidth]{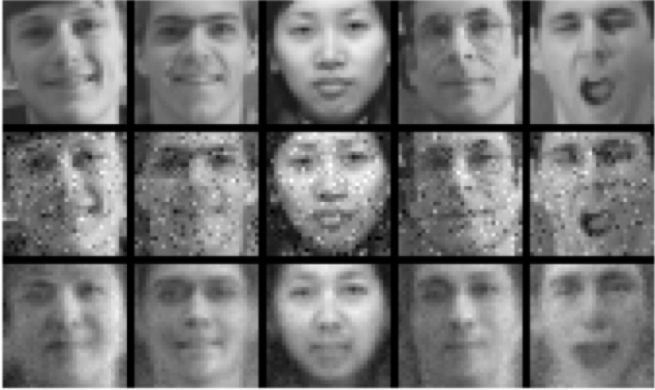}}\\
	\subfloat[]
	{
	\includegraphics[width=0.5\textwidth]{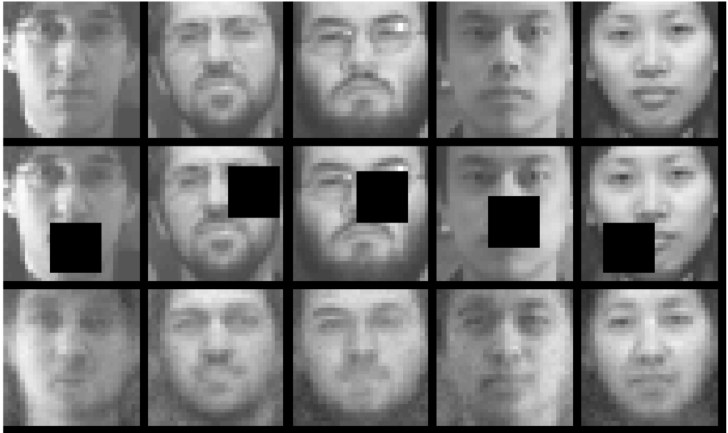}}
	\caption{Some examples of the face restoration from random noise (a) and block occlusion (b). Images collected from \cite{gross2010multi}. From top to bottom, the rows represent original images, corrupted images, and reconstructed images. }
	\label{fig:face}
\end{figure}

The second task for image representation is face restoration, where we use DRBN to restore a given cropped image. For this task,
we put two kinds of noise on the face images from Multi-PIE data set: random noise and block occlusion, 40\% of the pixels are corrupted by random noise with a standard deviation of 0.4, following the same procedure as~\cite{tang2012robust}. For the latter, 12$\times$12 blocks are superimposed on a random part of the 32$\times$32 faces with Gaussian random noise, following the same procedure as~\cite{tang2012robust}. The same MAP inference method, namely CA and AugCA is applied. Also, result from inference network (IN)~\cite{mnih2014neural} is compared with CA and AugCA. The baseline methods include the robust Boltzmann machine (RoBM)~\cite{tang2012robust}, Gaussian-Bernoulli RBM (GRBM)~\cite{hinton2006reducing}, and robust PCA (RPCA)~\cite{wright2009robust}. PSNR is employed to quantitatively evaluate different methods. The result is given in Table~\ref{facepsnr}. The AugCA method outperforms all the other methods in terms of PSNR values. In the cross-subject setting, RBN generalizes well to unseen subjects. Some examples of the reconstructed faces are given in Fig.~\ref{fig:face}.
\begin{table}
	\centering
	\caption{PSNRs for different denoising methods on Multi-PIE data set. }
	\scalebox{1.2}{
		\begin{tabular}{lcc}
			\toprule
			Method			& Random Noise	& Block Occlusion \\
			\midrule
			RPCA 			& 22.53			& 20.14 \\
			GRBM 			& 23.18			& 21.45 \\
			RoBM 			& 27.15			& 24.32 \\
			\midrule
			RBN (CA)      		& 28.60 	& 27.21 \\
			RBN (IN) \cite{mnih2014neural}     		& 27.89		& 26.21 \\
			RBN (AugCA)      & 29.23	 & 27.45 \\
			\bottomrule
		\end{tabular}
	}
	\label{facepsnr}
\end{table}

Next we demonstrate DRBN's image reconstruction ability. Different from the image restoration task presented in Fig. \ref{fig:face}, which restores images using trained DRBN models from a given corrupted image, for image reconstruction, given an image $\bm{x}$, we first perform MAP inference to obtain the most likely latent representation $\bm{h}^*$ and then obtain the reconstructed image $\bm{\hat x}$ via $\bm{\hat x}=\arg\max_{\bm{\hat x}}p(\bm{\hat x}|\bm{h}^*)$.

\begin{figure}[h]
	\centering
	\subfloat[]
	{
		\includegraphics[width=0.5\textwidth]{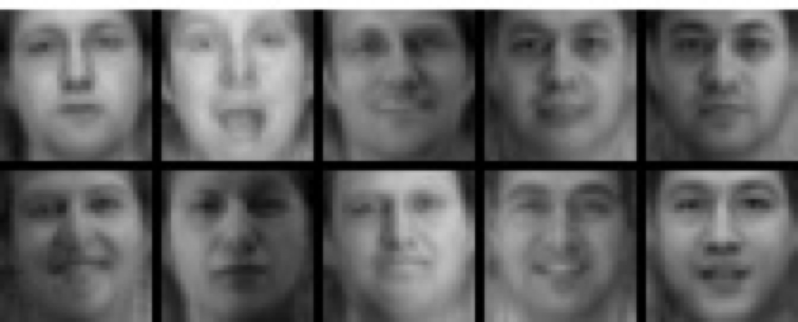} }\\
	\subfloat[]
	{
		\includegraphics[width=0.5\textwidth]{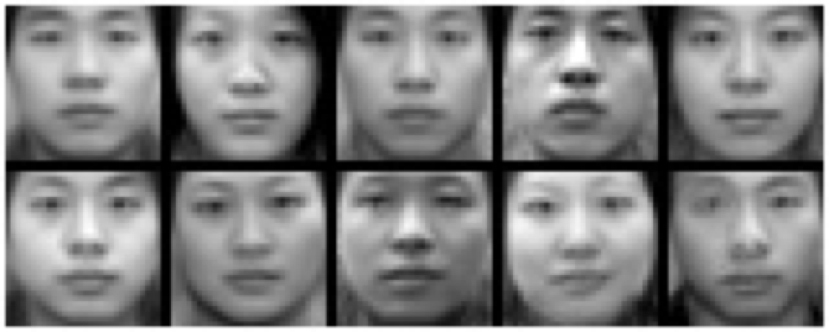} }\\
	
	\subfloat[]
	{
		\includegraphics[width=0.5\textwidth]{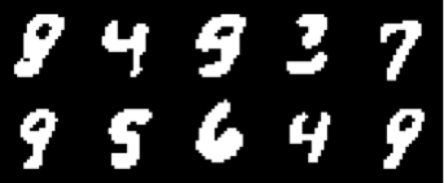} }\\
	\subfloat[]
	{
		\includegraphics[width=0.5\textwidth]{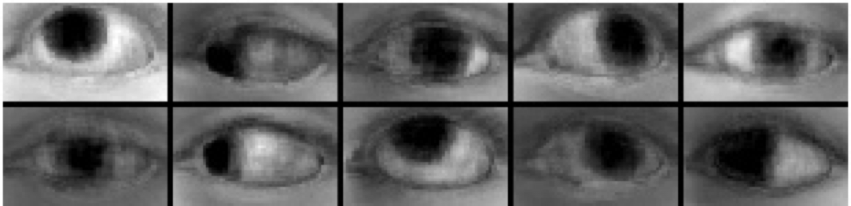} }
	\caption{Examples of the image reconstruction for (a) Multi-PIE, (b) CAS-PEAL, (c) MNIST, (d) UnityEye. }
	\label{fig:recon_image}
\end{figure}

Fig.~\ref{fig:recon_image} presents some examples. For Multi-PIE data set~\cite{gross2010multi}, we trained RBN with 200 latent variables, all face images are cropped and resized to 32$\times$32. Similarly, we trained different RBN on MNIST \cite{mnistdataset}, UnityEye~\cite{unityeye} and CAS-PEAL~\cite{caspeal} dataset, respectively. When training RBN on MNIST dataset, the number of hidden nodes is 50 for one hidden layer with the learning rate set to $0.01$. For both UnityEye and CAS-PEAL dataset, total hidden nodes are 200, image size are 60$\times$36 and 30$\times$30 accordingly. \ From the images  in Fig. \ref{fig:recon_image}
we can see that DRBN is able to effectively capture data distribution and can produce good image reconstruction.

\begin{figure}[h]
	\centering
	\subfloat[]
	{
	\includegraphics[width=0.5\textwidth]{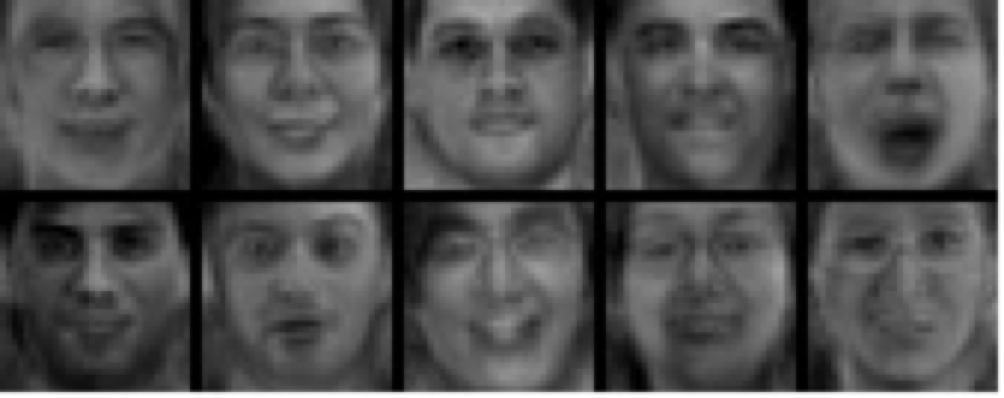} } \\
	\subfloat[]
	{
	\includegraphics[width=0.5\textwidth]{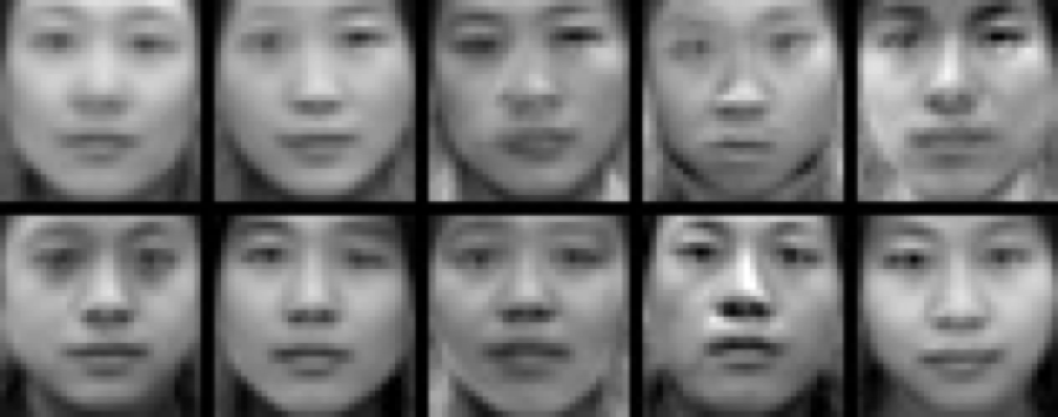} }
	\caption{An example of the image generation experiment. Generated images from (a) Multi-PIE, (b) CAS-PEAL }
	\label{fig:imagegen}
\end{figure}

The last task for assessing DRBN's data representation capability is through image generation. For this task, we use pretrained DRBN model in the image reconstruction task. To generate faithful images, we sample the latent variables from its prior probabilities. Given the values of the hidden nodes, images can be generated from sampling the conditional probability of visible nodes given latent variables. Examples of generated images from Multi-PIE and CAS-PEAL data sets are shown in Fig. \ref{fig:imagegen}.

\subsection{Feature Learning for Regression}
\label{sec:B}
In this section, we applied the discriminative supervised learning for feature learning using DRBN. Given the learnt DRBN model,
features in top layer can be extracted via MAP inference. The extracted features will then be used for regression tasks. We study head pose estimation and eye gaze estimation task with regression respectively.

The head pose estimation task is to predict three head pose angles (roll, pitch, and yaw) through a regression function. The Boston University head pose data set is used, the cropped faces are resized to 15$\times$15 pixels. Two latent layers are used, with 50 latent variables each. The MAP configuration of the top layer is obtained through both IN, CA and AugCA method, and is used as the features for a linear regression model. We compare the DRBN to three state-of-art model-based algorithms, namely 3D deformable head pose estimation (3D-Deform)~\cite{vicente2015driver}, deformable model fitting (DMF)~\cite{saragih2011deformable}, and monocular head pose estimation (MHPE)~\cite{morency2010monocular} and one learning-based algorithm, distance vector field with convolutional neural networks (DVF+CNN)~\cite{asteriadis2014visual}, with the mean absolute errors (MAEs) reported in Table~\ref{tab:pose}. The DRBN achieves comparable performance comparing to these competing algorithms. Note that the resolution of the face image used in DRBN is much smaller (15$\times$15) than other models (typically 320$\times$240), and no 3D information is used. This indicates the potential of using DRBN for head pose estimation in extremely low resolution images.
\begin{table}[h]
	\centering
	\caption{MAE of head pose angles in the BU data set.}%
	\scalebox{1.2}{
		\begin{tabular}{lcccc}
			\toprule
			Method & Yaw & Pitch & Roll\\
			\midrule
			DMF      & 5.2 & 4.5 & 2.6 \\
			3D-Deform    & 4.3 & 6.2 & 3.2 \\
			MHPE      & 5.0 & 3.7 & 2.9 \\
			DVF+CNN	  & 4.3 & 3.7 & 2.6 \\
			\midrule
			DRBN (IN) \cite{mnih2014neural} & 4.8 & 3.8 & 3.7 \\
			DRBN (CA) & 5.4 & 5.8 & 3.5 \\
			DRBN (AugCA) & 4.6 & 3.5 & 3.3 \\
			\bottomrule
		\end{tabular}
		\label{tab:pose}
	}
\end{table}

For the task of eye gaze estimation, we choose the MPII data set~\cite{zhang2015appearance} because it covers a wide range of recording locations and times, illuminations, and eye appearances. The data set contains 213,659 images from 15 subjects. The goal of our experiment is to map the eye images to the pitch and yaw eye gaze angles. The size of the cropped eye image is 36$\times$60, yielding a 2,160-dimension vector. The DRBN has a 2160-200-200 structure. To evaluate different models, we use the within-dataset leave-one-subject-out evaluation, in which the eye images from one subject are used for testing, and all the other images are used for training. The regression accuracies of different inference algorithms for each angle are given in Table~\ref{tab:gaze}. As competing algorithms, we include the convolutional neural network (CNN)~\cite{zhang2015appearance}, k-Nearest Neighbors (kNN)~\cite{sugano2014learning}, adaptive linear regression (ALR)~\cite{lu2014adaptive}, and support vector regression (SVR)~\cite{schneider2014manifold}. The DRBN achieves comparable performance in terms of the mean average errors, demonstrating its effectiveness to capture the inherent patterns in the eye images. The experiment results show that DRBNs are able to learn high quality features for regression tasks.

Experimental results in sections \ref{sec:A} and \ref{sec:B} show that DRBN can perform both generative and discriminative tasks, even though its discriminative performance may not match state of the art CNN models.
\begin{table}[h]
	\centering
	\caption{Regression errors for eye gaze estimation in the MPII data set.}
	\scalebox{1.2}{
		\begin{tabular}{lcccc}
			\toprule
			Method & Yaw & Pitch & MAE & Std.\\
			\midrule
			SVR \cite{schneider2014manifold}    & - & - & 6.6 & 0.6\\
			ALR  \cite{lu2014adaptive}  & - & - & 7.9 & 1.0\\
			kNN  \cite{sugano2014learning}  & - & - & 7.2 & 0.8\\
			CNN \cite{zhang2015appearance}   & - & - & 6.3 & 1.0\\
			\midrule
			DRBN (IN) 		& 5.8 & 3.7 & 7.6 & 1.1\\
			DRBN (CA) 		& 5.4 & 3.8 & 7.8 & 1.3 \\
			DRBN (AugCA) 	& 4.9 & 3.6 & 7.1 & 1.2 \\
			\bottomrule
		\end{tabular}
	}
	\label{tab:gaze}
\end{table}

\section{Summary}
In this paper, we review different structures of deep directed generative models and their learning and inference algorithms. We focus on a specific version of deep generative models, Deep Regression Bayesian Networks (DRBN). Compared to other deep learning models, DRBN can better capture the data distribution because of its explicitly capturing the dependencies among the latent variables.
Various algorithms are reviewed and compared, including the efficient inference and learning methods of replacing the max-sum operation with max-max operation
and the augmented coordinate ascent method for MAP inference.
 Extensive experiments on benchmark datasets for both data generation and classification tasks are presented, including image restoration, face reconstruction, head pose estimation, and eye gaze estimation.  They experiments demonstrate the competitive performance of DRBNs for both data representation and feature learning tasks.

\small
\bibliographystyle{plain}
\bibliography{bib}


\end{document}